\documentclass[letterpaper]{article}

\usepackage{aaai2026}
\usepackage{times}
\usepackage{helvet}
\usepackage{courier}
\usepackage[hyphens]{url}
\usepackage{graphicx}
\usepackage{natbib}
\usepackage{caption}
\usepackage{amsmath}
\usepackage{algorithm}
\usepackage{algpseudocode}
\usepackage{diagbox}
\usepackage{booktabs}
\usepackage{amssymb}
\usepackage{graphicx}
\usepackage{subcaption}
\usepackage{enumitem}
\usepackage{xcolor}
\usepackage{xspace}

\newcommand{\N}{\text{GeoGen}\xspace}

\title{GeoGen: A Two-stage Coarse-to-Fine Framework for Fine-grained \\Synthetic Location-based Social Network Trajectory Generation}

\author{
Rongchao Xu\textsuperscript{\rm 1},
Kunlin Cai\textsuperscript{\rm 2},
Lin Jiang\textsuperscript{\rm 1},
Zhiqing Hong\textsuperscript{\rm 3},
Yuan Tian\textsuperscript{\rm 2},
Guang Wang\textsuperscript{\rm 1}\thanks{Prof. Guang Wang is the corresponding author.}
}
\affiliations{
\textsuperscript{\rm 1}Florida State University, Tallahassee, Florida\\
\textsuperscript{\rm 2} University of California, Los Angeles, Los Angeles, California\\
\textsuperscript{\rm 3} Rutgers University, New Brunswick, New Jersey\\
rx21a@fsu.edu, kunlin96@g.ucla.edu, lin.jiang@fsu.edu, zh252@cs.rutgers.edu, yuant@ucla.edu, guang@cs.fsu.edu \\
}

\begin{document}

\maketitle

\begin{abstract}
Location-Based Social Network (LBSN) check-in trajectory data are important for many practical applications like POI recommendation, advertising, and pandemic intervention. However, the high collection costs and ever-increasing privacy concerns prevent us from accessing large-scale LBSN trajectory data. The recent advances in synthetic data generation provide us with a new opportunity to achieve this, which utilizes generative AI to generate synthetic data that preserves the characteristics of real data while ensuring privacy protection.
However, generating synthetic LBSN check-in trajectories remains challenging due to their spatially discrete, temporally irregular nature and the complex spatio-temporal patterns caused by sparse activities and uncertain human mobility.
To address this challenge, we propose GeoGen, a two-stage coarse-to-fine framework for large-scale LBSN check-in trajectory generation.
In the first stage, we reconstruct spatially continuous, temporally regular latent movement sequences from the original LBSN check-in trajectories and then design a Sparsity-aware Spatio-temporal Diffusion model (S$^2$TDiff) with an efficient denosing network to learn their underlying behavioral patterns.
In the second stage, we design Coarse2FineNet, a Transformer-based Seq2Seq architecture equipped with a dynamic context fusion mechanism in the encoder and a multi-task hybrid-head decoder, which generates fine-grained LBSN trajectories based on coarse-grained latent movement sequences by modeling semantic relevance and behavioral uncertainty.
Extensive experiments on four real-world datasets show that GeoGen excels state-of-the-art models for both fidelity and utility evaluation, e.g., it increases over 69\% and 55\% in distance and radius metrics on the FS-TKY dataset.

\end{abstract}

\begin{links}
\link{Code}{https://github.com/Rongchao98/GeoGen}
\end{links}

\section{Introduction}
Fine-grained location-based social network (LBSN) check-in trajectories are essential for many real-world applications such as next Point of Interest (POI)  recommendation~\cite{next_poi_rec_3}, urban mobility understanding \cite{wang2019urban}, business location selection \cite{xie2016learning}, advertising \cite{jeon2021lightmove}, and pandemic intervention \cite{hao2020understanding}. 
However, accessing large-scale LBSN data has become increasingly challenging due to high collection/purchasing costs and growing privacy concerns.  
Therefore, to support large-scale LBSN data access, it is crucial to develop methods that can generate synthetic LBSN check-in trajectories that maintain high fidelity and utility while safeguarding user privacy. 

\begin{figure}[t]
  \centering
  \includegraphics[width=\linewidth]{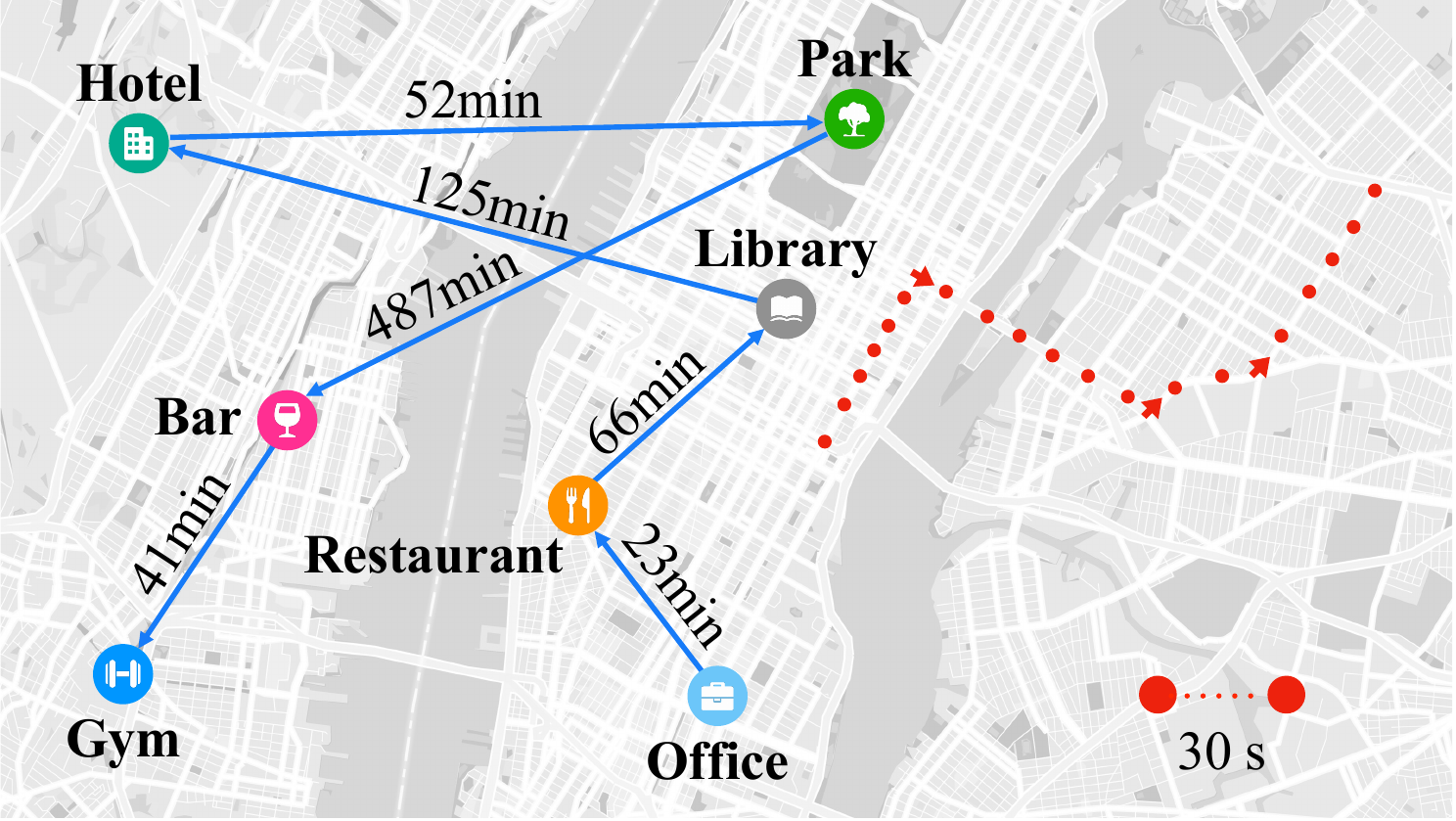}
  \caption{Example of a discrete LBSN trajectory with irregular intervals (blue); and a continuous GPS trajectory with a fixed interval of 30 seconds (red).}
  \label{asynchronous}
\end{figure}

To date, some methods have been proposed~\cite{isaacman2012human, jiang2016timegeo, yin2017generative, ouyang2018non, zhu2023difftraj} to generate LBSN check-in trajectory data. 
For instance, Generative adversarial networks (GANs)~\cite{ouyang2018non, feng2020learning, yu2017seqgan, yuan2022activity} have demonstrated substantial advances in generating LBSN check-in trajectories.
However, these approaches inevitably suffer from an unstable and extremely long training procedure.
Recently, the successes of diffusion models across modalities, including image \cite{dhariwal2021diffusion, kulikov2023sinddm} and audio \cite{huang2022prodiff} generation, have spurred investigations into trajectory generation.
Several diffusion-based models \cite{zhu2023difftraj, zhu2024controltraj, liu2023pristi} have been proposed to generate spatially continuous, temporally regular GPS trajectories with fixed intervals between consecutive data points and fixed lengths (e.g., 100 data points per trajectory). These models have demonstrated strong performance, highlighting the great potential of diffusion models for generating LBSN check-in trajectories.

However, it is nontrivial to adapt diffusion models to LBSN check-in trajectory data generation due to its two unique natures: 
(i) \textbf{Temporally irregular}: As shown in Figure~\ref{asynchronous}, in contrast to regularly sampled GPS trajectories, LBSN check-in trajectories exhibit complex spatio-temporal patterns due to uncertain human mobility and check-in behavior, leading to \textit{variable time intervals} between points and \textit{unfixed trajectory lengths}.
(ii) \textbf{Spatially discrete}: Diffusion models are well suited to \textit{fixed-length} data in the \textit{continuous} domain, whereas POIs lie in a discrete spatial space. Consequently, adapting models designed for regularly sampled trajectory generation \cite{zhu2023difftraj, zhu2024controltraj, liu2023pristi} to LBSN check-in trajectories is challenging: a large sampling interval (e.g., 60 minutes) degrades fidelity, while a small interval (e.g., 1 minute) severely reduces efficiency.

To address these challenges, we propose GeoGen, a two-stage, coarse-to-fine framework for synthetic LBSN trajectory data generation.
In the first stage, to adapt diffusion models to irregularly-sampled LBSN trajectory data, we reconstruct spatially continuous and temporally regular latent movement sequences from LBSN check-in trajectories through interpolation with a pre-determined coarse-grained interval.
We then design a Sparsity-aware Spatio-temporal Diffusion model (S$^2$TDiff), which includes a novel and efficient Spatially-Aware Sparsely-Gated U-Net (SASG-UNet) as its denoising network. 
SASG-UNet replaces standard convolutions with hierarchical 1D blocks to capture multi-scale patterns and integrates a specialized \textit{S$^2$G Attention} module to fit the unique characteristics of the latent movement sequences.    
In the second stage, we propose a Transformer-based Seq2Seq architecture {Coarse2FineNet} that auto-regressively generates fine-grained check-in points based on the coarse-grained latent movement sequences.
Coarse2FineNet incorporates a POI Context-aware Encoder that captures rich contextual representations of the latent movement sequence, and a Multi-task Hybrid-Head Decoder that generates the next POI and its fine-grained timestamp.
In the encoder, we introduce a \textit{dynamic context fusion} mechanism that aligns the input latent movement sequence with POIs exhibiting diverse spatio-temporal characteristics, integrating them into unified contextual representations.  
In the decoder, instead of directly predicting the next check-in point, we leverage a temporal point process to model the user behavior uncertainty underlying LBSN check-in trajectories.
This two-stage approach effectively resolves a key trade-off between computational efficiency and data quality, enabling the generation of large-scale synthetic LBSN check-in trajectories.

The key contributions of this paper are as follows:
\begin{itemize}
    \item Conceptually, we focus on developing innovative diffusion models to efficiently generate large-scale fine-grained LBSN check-in trajectories, which are inherently spatially discrete and temporally irregular.

    \item Technically, we propose a two-stage coarse-to-fine framework called GeoGen for synthetic LBSN trajectory generation. In the first stage, we design a novel S$^2$TDiff model with an efficient SASG-UNet as a denoising network to generate coarse-grained latent movement sequences. In the second stage, we design a Transformer-based Seq2Seq architecture, Coarse2FineNet, which includes a POI Context-aware Encoder and a Multi-task Hybrid-head Decoder to generate fine-grained check-ins at precise, non-uniform timestamps.

    \item Experimentally, we conduct extensive experiments on four real-world LBSN check-in datasets. The results demonstrate that GeoGen significantly outperforms state-of-the-art models in data fidelity and show high utility for downstream tasks like next location prediction. 
\end{itemize}

\section{Preliminary}

\subsection{Problem Statement}
\textbf{Definition 1. (LBSN Check-in Trajectories).} 
An LBSN check-in trajectory is a sequence of POI visits from an individual, denoted $s = [x_1, x_2, \ldots]$.
The \(i\)th element \(x_i\) is a check-in represented as the tuple \(x_i=(p_i,t_i)\), where \(p_i\in P\) is a POI ID from a local finite set \(P\) with associated geographic coordinate \(g_{p_i}\), and \(t_i\) is the check-in timestamp.

\noindent \textbf{Definition 2. (Trajectory Granularity)}
The \textbf{granularity} of a trajectory is its smallest recorded time unit. \textit{Coarse-grained trajectories} use large time units (e.g., hours), while \textit{fine-grained trajectories} use smaller ones (e.g., minutes).
This work aims to generate fine-grained LBSN check-in trajectories with a minute-level granularity.

\noindent \textbf{Definition 3. (LBSN Check-in Trajectory Generation).} \label{def-lbsn_trajs} 
Given a real-world dataset containing fine-grained LBSN check-in trajectories, denoted as $S = [s_1, s_2, \ldots]$, the LBSN check-in trajectory generation task aims to generate a synthetic trajectory dataset $\hat{S} = [\hat{s}_1, \hat{s}_2, \ldots]$ that maintains the spatio-temporal characteristics and distributions of $S$, as well as high data utility for downstream applications. 

\begin{figure*}[t]
    \centering
    \includegraphics[width=0.99\linewidth]{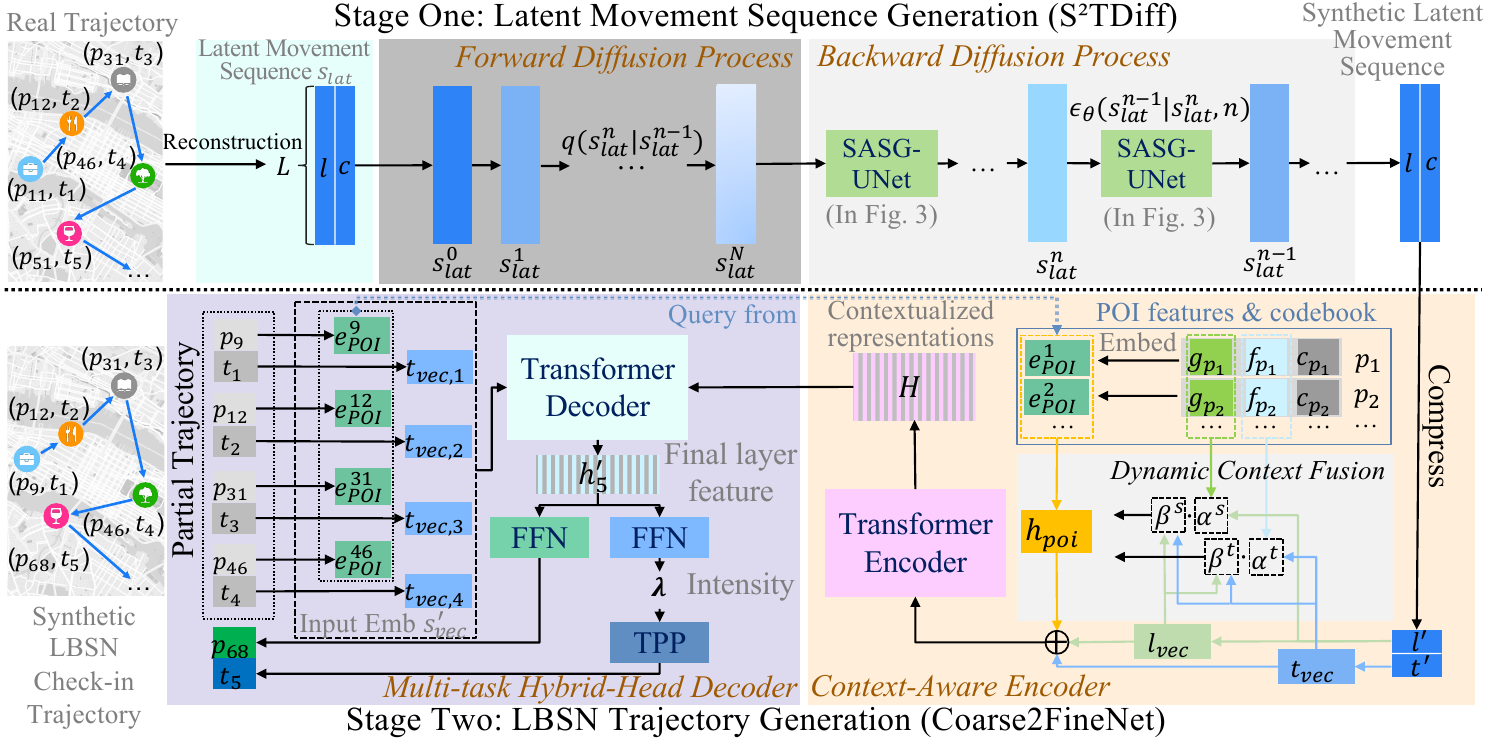}
    \caption{The overall framework of our proposed GeoGen. In the first stage, within S$^2$TDiff, each reconstructed latent movement sequence $s_{\text{lat}}$ is first transformed into pure noise $s_{\text{lat}}^N$. During inference, SASG-UNet progressively denoises a sampled noise sequence to generate a synthetic latent trajectory. In the second stage, the context-aware encoder of Coarse2FineNet extracts contextualized representations from the generated latent movement sequence. The decoder then auto-regressively generates each POI and fine-grained timestamp pair, conditioned on both the contextual representations and the partially generated LBSN check-in trajectory. FFN and TPP refer to the feed-forward network and temporal point process, respectively.}
    \label{fig:framework}
\end{figure*}

\subsection{Denoising Diffusion Probabilistic Model}
\label{DPM}
Denoising Diffusion Probabilistic Models \cite{ho2020denoising} encompass two processes: (i) a forward diffusion process that gradually introduces noise into the data, and (ii) a backward diffusion process that reconstructs the original data from its noisy state.
The forward diffusion process is a non-trainable Markov chain that transforms real sample $\mathbf{x}^0$ into latent variables $\mathbf{x}^1, \ldots, \mathbf{x}^N$, which can be represented as: 
\begin{equation}
    q(\mathbf{x}^n|\mathbf{x}^{n-1}) = \mathcal{N}(\mathbf{x}^n; \sqrt{1 - \beta_n}\mathbf{x}^{n-1}, \beta_n \mathbf{I}),
\end{equation}
where $\beta_1, \ldots, \beta_N \in (0, 1)$ are pre-determined variance schedules. $\mathbf{x}^n$ can be directly sampled from $\mathbf{x}^0$ with $q(\mathbf{x}^n|\mathbf{x}^0) = \mathcal{N}(\mathbf{x}^n; \sqrt{\overline{\alpha}_n}\mathbf{x}^0, (1 - \overline{\alpha}_n)\mathbf{I})$, where $\alpha_n = 1 - \beta_n$ and $\overline{\alpha}_n = \prod_{i=1}^N \alpha_i$. With the application of the reparameterization, $\mathbf{x}^n$ can be represented as $\mathbf{x}^n = \sqrt{\overline{\alpha}_n}\mathbf{x}^0 + \sqrt{1 - \overline{\alpha}_n}\boldsymbol{\epsilon}$, where $\boldsymbol{\epsilon} \sim \mathcal{N}(0, \mathbf{I})$.
The backward diffusion process is a trainable Markov chain that aims to recover $\mathbf{x}^0$ from $\mathbf{x}^N$, which can be formulated as:
\begin{equation}
    p_{\theta}(\mathbf{x}^{n-1}|\mathbf{x}^n) =
\mathcal{N}(\mathbf{x}^{n-1}; \mu_{\theta}(\mathbf{x}^n, n), \sigma_{\theta}^2(\mathbf{x}^n, n)\mathbf{I}),
\end{equation}
where $\mu_{\theta}(\mathbf{x}^n, n)$ and $\sigma_{\theta}(\mathbf{x}^n, n)$ are the mean and variance predicted normally by a neural network parameterized by $\theta$. 
The loss function is formalized as:

\begin{equation} 
\mathbb{E}_{\mathbf{x}^0,\boldsymbol{\epsilon},\,n}\!\left[
\left\|\boldsymbol{\epsilon}-\epsilon_{\theta}\!\left(\sqrt{\overline{\alpha}_n}\mathbf{x}^0+\sqrt{1-\overline{\alpha}_n}\,\boldsymbol{\epsilon},\,n\right)\right\|^2
\right].
\end{equation}
where $\epsilon_{\theta}$ is a neural network for predicting sampled $\boldsymbol{\epsilon} \sim \mathcal{N}(0, \mathbf{I})$. After training, trajectory generation is conducted by progressively sampling $\mathbf{x}^{n-1}$ from distribution $p_{\theta}(\mathbf{x}^{n-1}|\mathbf{x}^n)$ until reach $\mathbf{x}^{0}$ by computing:
\begin{equation}
    \mathbf{x}^{n-1}
= \frac{1}{\sqrt{\alpha_n}}\!\left(
    \mathbf{x}^n - \frac{\beta_n}{\sqrt{1-\overline{\alpha}_n}}\,
    \epsilon_{\theta}(\mathbf{x}^n, n)
  \right)
  + \sigma_{\theta}(\mathbf{x}^n,n)\,\mathbf{z},
\end{equation}
where \( \mathbf{z} \sim \mathcal{N}(\mathbf{0}, \mathbf{I}) \) for \( n \in [2, N] \), and \( \mathbf{z} = \mathbf{0} \) when \( n = 1 \).

\section{Methodology}

In this part, we introduce the detailed design of \N. An overall pipeline of \N is shown in Figure~\ref{fig:framework}.

\subsection{Coarse-grained Latent Movement Sequence Generation}
\label{coarse-grained-trajGen}
To adapt diffusion models to irregularly-sampled LBSN trajectory data, we first convert original fine-grained LBSN trajectories into coarse-grained spatially continuous and temporally regular latent movement sequences. We then design a Sparsity-aware Spatio-temporal Diffusion model (S$^2$TDiff) to generate synthetic coarse-grained latent movement sequences by learning underlying spatio-temporal distributions and physical constraints.

\subsubsection{Coarse-grained Latent Movement Sequence Reconstruction}
\label{reconstruction}
Given a time interval $\mathit{I}$, we divide the total duration into $L$ slots, where $L = [D / I]$, $D$ is the duration.
We reconstruct each real LBSN check-in trajectory $s$ into a latent movement sequence, $s_{\mathit{lat}} = [(l_1, c_1), \ldots, (l_L, c_L)]$. 
For each slot $i$, the latent coordinate $l_i$ is the geographic mean of all check-ins $P_i$ falling within that time window ($l_i = \frac{1}{|P_i|}\sum_{(p_j,t_j) \in P_i} g_{p_j}$), and the intensity $c_i$ is their count ($c_i = |P_i|$).
Next, we adopt \textit{linear interpolation} between the nearest preceding ($i_p$) and subsequent ($i_n$) time slots with no check-ins ($c_i = 0$), where the missing location is calculated as $l_i = l_{i_p} + ((i - i_p) / (i_n - i_p)) \cdot (l_{i_n} - l_{i_p})$.
For empty slots at the sequence boundaries, we apply \textit{circular interpolation}, treating the trajectory as a loop to infer locations between the first and last observed points.
This reformulation transforms sparse event data into a \textbf{\textit{spatially continuous}} and \textbf{\textit{temporally regular}} format, enabling compatibility with diffusion models while reducing memory and computational costs via temporal compression.

\subsubsection{Sparsity-aware Spatio-temporal Diffusion Model}
\begin{figure}[htbp]
    \centering
    \includegraphics[width=\linewidth]{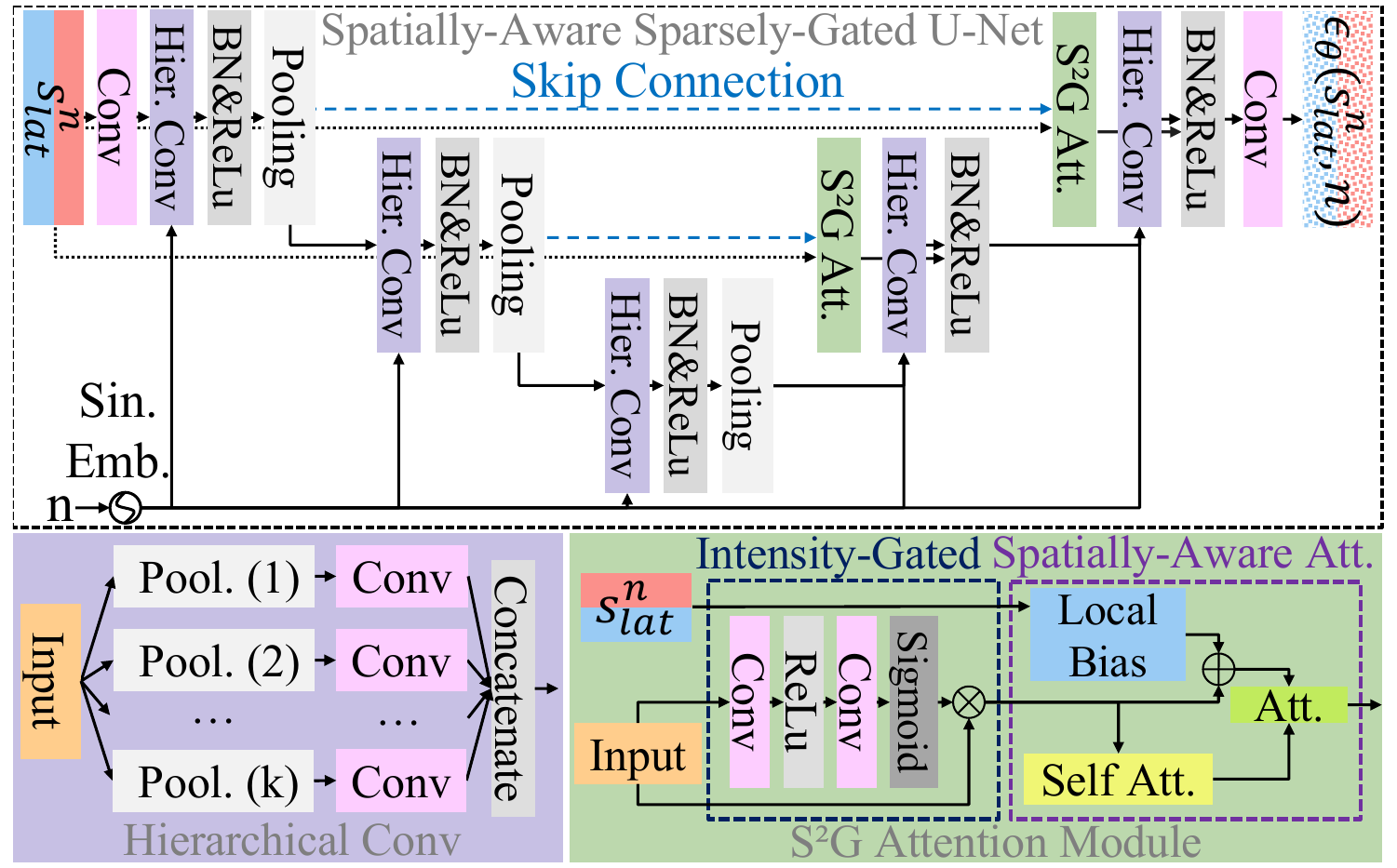}
    \caption{The architecture of the proposed Spatially-Aware Sparsely-Gated U-Net (SASG-UNet). The kernel sizes for the average pooling layers in the Hierarchical 1D Convolution layers are indicated in parentheses. }  
    \label{fig:framework}
\end{figure}

To capture the underlying spatio-temporal patterns, we propose a Sparsity-aware Spatio-temporal Diffusion model (S$^2$TDiff), trained with latent movement sequences $S_{\text{lat}} = [s_{\text{lat},1}, s_{\text{lat},2}, \dots]$ as input. 
Although linear interpolation yields spatially continuous sequences, it poses challenges of (1) \textit{diverse movement rates} and (2) \textit{sparse intensity}. 


To address these challenges, we design a novel and efficient \textit{Spatially-Aware Sparsely-Gated U-Net} (SASG-UNet) as the denoising network in S$^2$TDiff.
S$^2$TDiff employs a U-Net backbone with hierarchical 1D convolution blocks that replace standard 2D convolutions. These blocks efficiently capture multi-resolution features using parallel branches of multi-scale average pooling and fixed-size 1D convolutions. At training/inference step $n$, the denoising network SASG-UNet $\epsilon_{\theta}$ takes a noisy latent movement sequence $s_{\text{lat}}^n$ as input and outputs the predicted noise: $\epsilon_{\theta}(s_{\text{lat}}^n, n)$.
To further tackle sparse intensity and diverse movement rates, we introduce a novel S$^2$G Attention module within SASG-UNet’s skip connections, consisting of two complementary pathways: the \textbf{\textit{Intensity-Gated Pathway}} and the \textbf{\textit{Spatially-Aware Attention Pathway}}.
Firstly, the Intensity-Gated Pathway handles sparse signals by with a lightweight convolutional sub-network to generate a dynamic gating signal \(g = \sigma(\text{Conv1D}(\text{ReLU}(\text{Conv1D}(h_{\text{enc}}))))\), 
where $h_{\text{enc}}$ denotes encoder features and $\sigma$ is a sigmoid function.
This gate adaptively modulates the features, allowing the network to prioritize learning from informative, sparse, non-zero intensity values.
Secondly, the Spatially-Aware Attention Pathway accounts for diverse movement rates via a \textit{Local Spatial Bias} $b$, where at index $i$, \(b_i = \mathcal{F}_{\text{emb}}(d_{\text{H}}((s^n_{\text{lat}, i}, s^n_{\text{lat}, i-1}))\), and $d_{\text{H}}$ is the Haversine distance~\cite{maria2020measure} between consecutive points $s^n_{\text{lat}, i}$, $ s^n_{\text{lat}, i-1}$ in the noisy input and $\mathcal{F}_{\text{emb}}$ is a learnable feed-forward network.  
Next, we add an attention mechanism, yielding the final refined features:
\(h_{\text{att}} = \text{softmax}\left(\frac{QK^T}{\sqrt{d_k}}\right)V\),
where $Q$ and $K$ are the self-attention output features of the Intensity-Gated Pathway and $V$ is the bias $b$.
This mechanism provides the model with a direct and efficient understanding of real-world movement dynamics, enabling it to handle sparse, irregular mobility patterns effectively.

\subsection{Coarse2FineNet for Fine-grained LBSN Trajectory Generation with Context Awareness}
\label{Coarse2Fine}
In the second stage, we propose \textbf{Coarse2FineNet}, a Transformer-based Seq2Seq framework that translates coarse-grained latent movement sequences generated by S$^2$TDiff into fine-grained LBSN check-in trajectories by jointly modeling the spatio-temporal features of the latent movement sequence and characteristics of POIs. 
Coarse2FineNet includes two key components: a Context-Aware Encoder and a Multi-task Hybrid-Head Decoder. 
The encoder captures a rich, contextualized representation of user mobility from coarse-grained latent movement sequences, while the decoder autoregressively generates a sequence of POI visits with fine-grained timestamps.

\subsubsection{Context-Aware Encoder}
To reduce complexity in the diffusion process, we compress a latent sequence $s_{\text{latent}} = \{(l_i, c_i)\}$ into $s_{\text{latent}}' = \{(l'_i, t'_i) \mid c_i \geq \gamma\}$ by filtering points, where $l'_j = l_{i_j}$ and 
$t'_j = t_{i_j} = (i_j + 0.5) \cdot \mathit{I}$
, $\gamma$ denotes intensity threshold, and $t_i$ denotes the midpoint time of slot $i$. 
We first obtain spatial embeddings $l_{vec} = Linear(l'_i) $ and temporal embedding $t_{vec} = T2V(t'_i)$, where $T2V$ denotes the Time2Vec method ~\cite{kazemi2019time2vec} that captures both linear and cyclical temporal patterns.

To establish local POI context awareness, we design a \textbf{\textit{Dynamic Context Fusion}} (DCF) module within the encoder.
In DCF, a shared POI embedding codebook is built to encode multiple attributes, including geographic coordinates $g_{p_j}$, temporal visit frequency vectors $f_{p_j}$, and category one-hot vector $c_{p_j}$ for POI ${p_j}$:
\(e_{\text{POI}}^{(j)} = W_{\text{latlon}} g_{p_j} + W_{\text{freq}} f_{p_j} + W_{\text{cat}} c_{p_j}\),
where $W_{\text{latlon}}, W_{\text{freq}}, W_{\text{cat}}$ are learnable weights.
Then, we employ a dual attention mechanism to model spatial and temporal relationships separately.
The spatial attention maps each coordinate $l'_i$ from the filtered sequence to a distribution over POIs based on proximity. 
The attention weight $\alpha_j$ for each POI $j$ and coordinate $l_i$ is calculated as:
\begin{equation}
    \alpha_{i,j}^{(s)} = \frac{\exp(-\|l'_i - g_{p_j}\| / \tau_s)}{\sum_{k=1}^{|\mathcal{P}|} \exp(-\|l'_i - g_{p_k}\| / \tau_s)}
\end{equation}
where $\tau_s$ is a temperature parameter that controls the distribution's sharpness. 
Concurrently, the temporal attention captures time-specific visit patterns. 
The temporal embedding $t'_{vec,i}$ is projected into a query vector $w_i = W_t t'_{vec,i}$, and the attention weight is computed via dot-product similarity:
\begin{equation}
    \alpha_{i,j}^{(t)} = \frac{\exp(f_{p_j} \cdot w_i)}{\sum_{k=1}^{|\mathcal{P}|} \exp(f_{p_k} \cdot w_i)}
\end{equation}
Finally, we adaptively balance the spatial and temporal signals by computing fusion weights using a softmax over their average representations \((\beta_s, \beta_t) = \text{Softmax} \left( \mathbf{W}_\alpha \left[ \frac{1}{L} \sum_{i=1}^{L} \mathbf{l}_i \,\|\, \frac{1}{L} \sum_{i=1}^{L} \mathbf{t}_i \right] \right)\)
, where $\mathbf{l}_{i}$ and $\mathbf{t}_{i}$ denote the $i$-th spatial and temporal feature vectors, respectively; $\mathbf{W}_\alpha$ is a learnable linear projection matrix that is used to create a unified attention distribution $\boldsymbol{\alpha}_{i,j} = \beta_s \boldsymbol{\alpha}^{(s)}_{i,j} + \beta_t \boldsymbol{\alpha}^{(t)}_{i,j}$.
This fused attention is applied to the shared POI embeddings to create a context-aware representation $h_{\text{poi},i} = \sum_{j=1}^{|\mathcal{P}|} \alpha_{i,j} e_{\text{POI}}^{(j)}$.
This POI-enhanced vector $h_{poi}$ is integrated with the initial features $h_{vec} = [l_{vec}, t_{vec}]$ through a residual connection, and is processed by a Transformer Encoder to produce the contextualized representations $H$.

\begin{table*}[t]
\centering
\small 
\resizebox{\textwidth}{!}{ 
\begin{tabular}{|c|c|c|c|c|c|c|c|c|c|c|}
\hline
\diagbox{Method}{Metric} & Distance & Radius & Interval & Length & Average & Distance & Radius & Interval & Length & Average \\ \hline \hline

\multicolumn{1}{|c|}{} & \multicolumn{5}{c|}{FS-NYC} & \multicolumn{5}{c|}{FS-TKY} \\ \hline
SMM~\cite{maglaras2015social} & 0.196 & 0.259 & 0.381 & 0.358 & 0.299 & 0.178 & 0.277 & 0.217 & 0.329 & 0.250 \\
TimeGEO~\cite{jiang2016timegeo} & 0.251 & 0.625 & 0.278 & 0.386 & 0.385 & 0.277 & 0.524 & 0.232 & \underline{0.216} & 0.312 \\
Hawkes~\cite{laub2015hawkes} & 0.190 & 0.517 & 0.339 & 0.253 & 0.325 & 0.157 & 0.315 & 0.269 & 0.281 & 0.256 \\
LSTM~\cite{rossi2021vehicle} & 0.166 & 0.441 & 0.175 & \underline{0.235} & 0.254 & 0.184 & 0.292 & 0.230 & 0.257 & 0.241 \\
SeqGAN~\cite{yu2017seqgan} & 0.116 & 0.364 & 0.150 & 0.291 & 0.230 & 0.108 & 0.292 & 0.207 & 0.276 & 0.221 \\
MoveSim~\cite{feng2020learning} & 0.080 & 0.289 & 0.133 & 0.391 & 0.223 & 0.079 & 0.256 & \underline{0.114} & 0.353 & 0.201 \\
DiffTraj~\cite{zhu2023difftraj} & \textbf{0.051} & \underline{0.085} & \underline{0.104} & 0.320 & \underline{0.140} & 0.276 & 0.201 & 0.116 & 0.744 & 0.334 \\
ControlTraj~\cite{zhu2024controltraj} & 0.193 & 0.356 & 0.160 & 0.337 & 0.262 & \underline{0.049} & \underline{0.096} & 0.191 & 0.223 & \underline{0.140} \\
\textbf{Our Method (ours)} & \underline{0.079} & \textbf{0.065} & \textbf{0.067} & \textbf{0.127} & \textbf{0.085} & \textbf{0.015} & \textbf{0.043} & \textbf{0.081} & \textbf{0.120} & \textbf{0.065} \\ \hline \hline

\multicolumn{1}{|c|}{} & \multicolumn{5}{c|}{FS-ATX} & \multicolumn{5}{c|}{GW-STO} \\ \hline
SMM~\cite{maglaras2015social} & 0.603 & 0.625 & 0.139 & 0.347 & 0.429 & 0.637 & 0.599 & 0.697 & 0.428 & 0.590 \\
TimeGEO~\cite{jiang2016timegeo} & 0.416 & 0.501 & 0.148 & 0.401 & 0.367 & 0.579 & 0.489 & 0.727 & 0.397 & 0.548 \\
Hawkes~\cite{laub2015hawkes} & 0.517 & 0.556 & 0.261 & 0.521 & 0.464 & 0.621 & 0.496 & 0.579 & 0.433 & 0.532 \\
LSTM~\cite{rossi2021vehicle} & 0.589 & 0.602 & 0.187 & 0.339 & 0.429 & 0.618 & 0.527 & 0.699 & 0.369 & 0.553 \\
SeqGAN~\cite{yu2017seqgan} & 0.452 & 0.475 & \textbf{0.029} & 0.292 & 0.312 & 0.531 & 0.462 & 0.745 & 0.413 & 0.538 \\
MoveSim~\cite{feng2020learning} & 0.306 & 0.382 & 0.078 & \underline{0.275} & 0.260 & 0.181 & 0.265 & \underline{0.318} & \underline{0.138} & \underline{0.226} \\
DiffTraj~\cite{zhu2023difftraj} & \underline{0.061} & \underline{0.104} & 0.241 & 0.702 & 0.277 & \textbf{0.114} & 0.275 & 0.802 & 0.306 & 0.374 \\
ControlTraj~\cite{zhu2024controltraj} & 0.108 & 0.217 & 0.150 & 0.286 & \underline{0.190} & 0.204 & \underline{0.253} & 0.765 & 0.141 & 0.341 \\
\textbf{Our Method (ours)} & \textbf{0.058} & \textbf{0.088} & \underline{0.036} & \textbf{0.200} & \textbf{0.096} & \underline{0.136} & \textbf{0.197} & \textbf{0.292} & \textbf{0.080} & \textbf{0.176} \\ \hline
\end{tabular}
}
\caption{Fidelity Evaluation. The best results on each dataset are in \textbf{bold}, and the second-best results are \underline{underlined}. `Average' denotes the mean value of the four metrics.}
\label{tab:fidelity}
\end{table*}

\subsubsection{Multi-task Hybrid-Head Decoder}
The Multi-task Hybrid-Head Decoder autoregressively generates a sequence of (POI, timestamp) pairs, conditioned on the contextual representations $H$ from the encoder.
To generate the $n'$-th check-in point, the previously generated sequence $s'_{\mathit{syn}} = [(p_1, t_1), \ldots, (p_{n'-1}, t_{n'-1})]$ is first embedded into a sequence of vectors $s'_{vec}$, where each POI is replaced by its shared embedding and each timestamp is converted using Time2Vec.
Next, a Transformer-based Decoder processes this embedded sequence and the encoder's output $H$ to produce a hidden state $h_{n'}$, which summarizes the necessary history and context.
This hidden state is then fed into two specialized prediction heads to determine \textit{where} and \textit{when} the next check-in occurs.
For POI, a linear layer computes logits over the vocabulary.
For timestamps, unlike direct regression, we adopt a \textit{Neural Temporal Point Process} (NTPP)~\cite{shchur2021neural} to capture the irregular timing of human mobility. 
The NTPP models the probability distribution of the next timestamp through a conditional intensity function
\(\lambda(t) = \text{Softplus}(W_{\text{time}} h_{n'} + b_{\text{time}}),\)
which defines the instantaneous event rate given the current context.
To sample the next timestamp, we first derive the cumulative distribution function \(F^{-1}\) corresponding to $\lambda(t)$, and then draw a sample from this distribution via inverse transform sampling:
\begin{equation}
t_{n'} = F^{-1}(u), \quad \text{where } u \sim \mathcal{U}(0,1).
\end{equation}
This enables the model to generate diverse and realistic non-uniform time intervals, capturing the stochastic nature of real-world check-in behaviors.

\subsubsection{Training Objective}
Coarse2FineNet is trained with teacher forcing and causal masking \cite{vaswani2017attention}.
Our training objective 
is formulated as:
\begin{equation}
    \mathcal{L} = \lambda_{\text{POI}} \mathcal{L}_{\text{POI}} + \lambda_{\text{time}} \mathcal{L}_{\text{time}} + \lambda_{\text{spatial}} \mathcal{L}_{\text{spatial}}.
\end{equation}
The POI prediction loss $\mathcal{L}_{\text{POI}} = -\sum_{t=1}^T y_t \log(\hat{y}_t)$ enforces categorical accuracy through cross-entropy, where $T$ represents the number of check-ins in a trajectory and $y_t$, $\hat{y}_t$ denote ground-truth and predicted POI ID respectively. 
The temporal loss is the negative log-likelihood objective for the neural temporal point processes, formulated as
\(\mathcal{L}_{\text{time}} = -\sum_{i=1}^T (\log \lambda_i - \lambda_i \Delta t_i)\)
, where $\lambda_i$ is the intensity predicted by the network for the $i$-th interval, and $\Delta t_i$ is the ground-truth time duration since the last check-in. This loss maximizes the probability of the observed check-in sequence by training the model to predict a high intensity just before a check-in occurs.
The spatial consistency loss $\mathcal{L}_{\text{spatial}} = \frac{1}{T} \sum_{t=1}^T \|g_{p_t} - g_{p_{\hat{t}}}\|_2$ penalizes geographic implausibility by measuring Euclidean distance between predicted POI $p_{\hat{t}}$ and ground-truth POI $p_t$. The three components are balanced by three learnable weights, enabling the model to generate trajectories that are spatially plausible, temporally accurate, and categorically consistent.

\section{Evaluation}






\begin{figure*}[thbp]
\centering

\begin{subfigure}[b]{0.24\textwidth}
    \includegraphics[width=\linewidth]{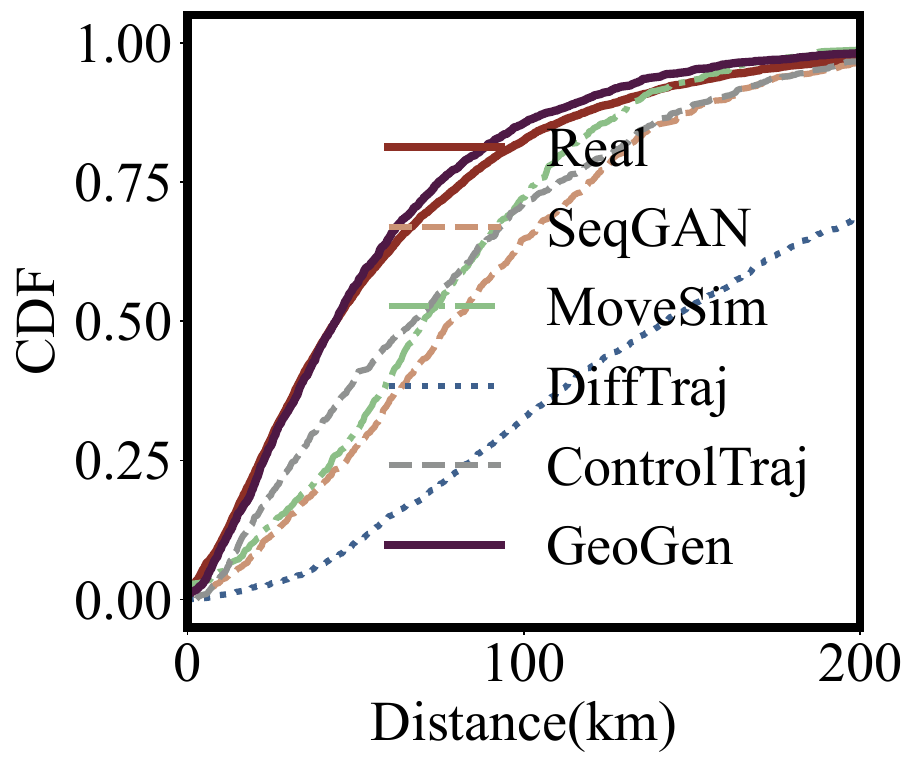}
    \caption{CDF of distance}
    \label{fig:sub1}
\end{subfigure}
\hfill
\begin{subfigure}[b]{0.24\textwidth}
    \includegraphics[width=\linewidth]{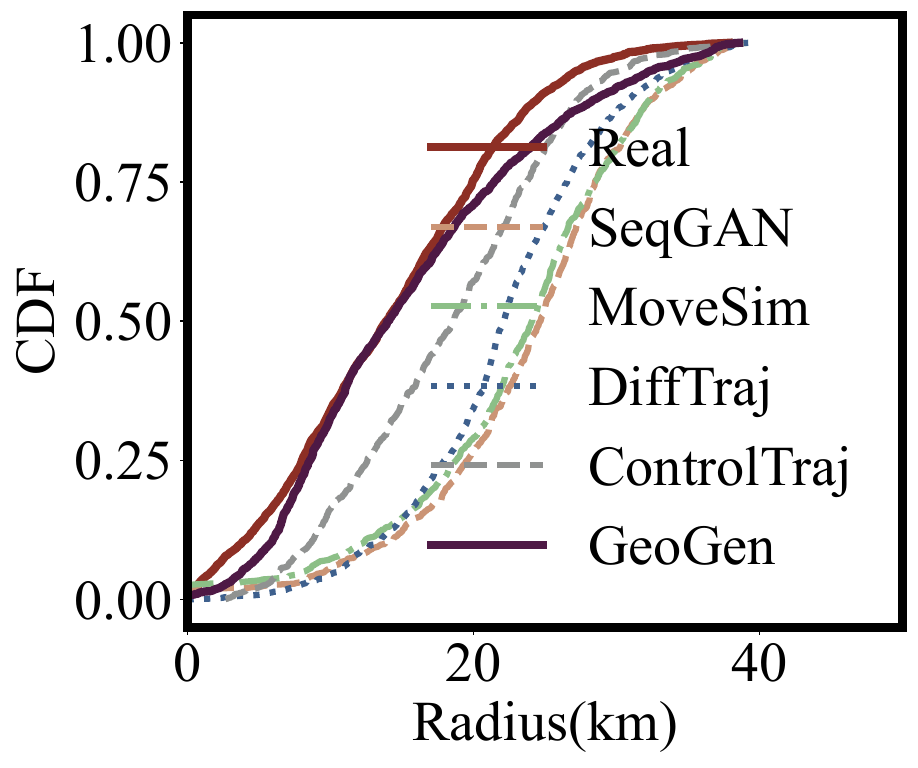}
    \caption{CDF of radius}
    \label{fig:sub2}
\end{subfigure}
\hfill
\begin{subfigure}[b]{0.24\textwidth}
    \includegraphics[width=\linewidth]{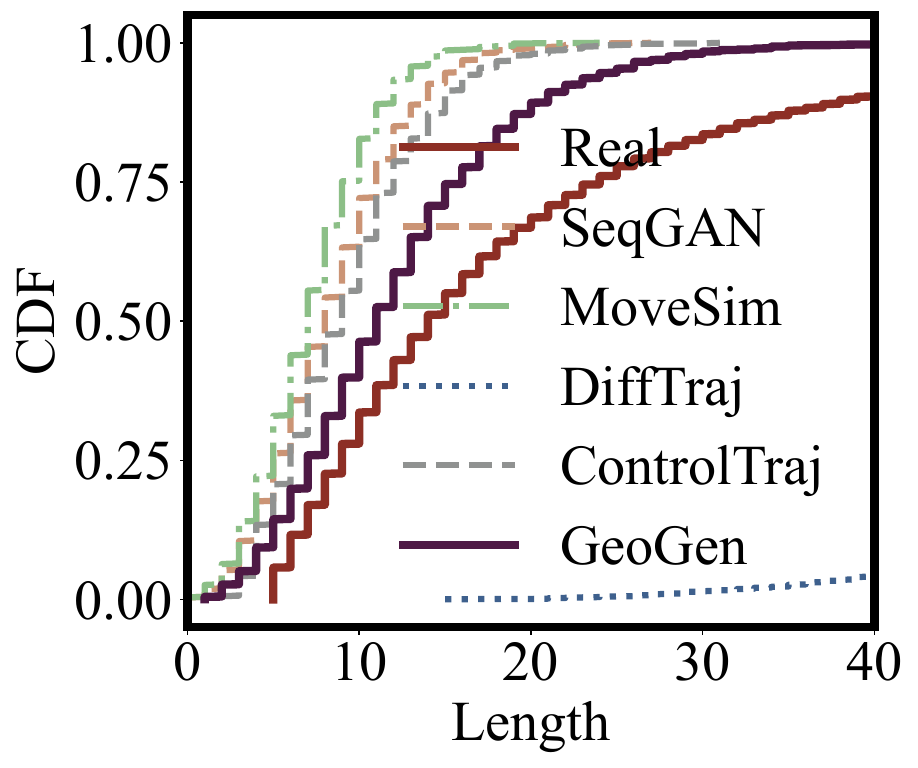}
    \caption{CDF of length}
    \label{fig:sub3}
\end{subfigure}
\hfill
\begin{subfigure}[b]{0.24\textwidth}
    \includegraphics[width=\linewidth]{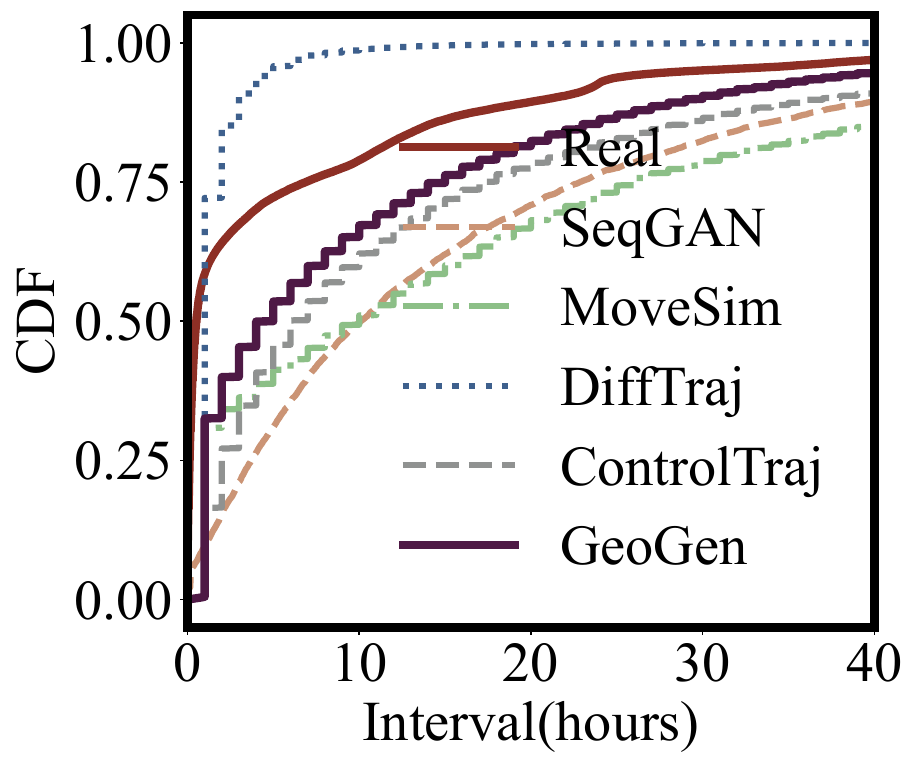}
    \caption{CDF of interval}
    \label{fig:sub4}
\end{subfigure}

\caption{Performance comparison of different methods on FS-TKY dataset.}
\label{fig:fidelity}
\end{figure*}

\subsection{Experiment Setup}\label{Setup}
\subsubsection{Datasets}
We utilize four publicly available LBSN datasets containing POI check-in trajectories to evaluate the performance of our proposed \N, including New York City (FS-NYC), Tokyo (FS-TKY), Austin (FS-ATX) from the Foursquare dataset\cite{yang2014modeling, yang2016participatory}, and Stockholm (GW-STO) from the Gowalla dataset \cite{cho2011friendship}. We first truncate each individual LBSN check-in trajectory with a pre-defined duration (one week for the three FourSquare datasets and six weeks for GW-STO) and maintain the sequences with more than 5 visits to ensure data quality. 
We split each dataset into training, validation, and testing sets with a ratio of 7:1:2.

\subsubsection{Baselines}
We compare our \N\ with the following eight state-of-the-art models: 
SMM \cite{maglaras2015social}, 
TimeGEO \cite{jiang2016timegeo},
Hawkes \cite{laub2015hawkes},
LSTM \cite{rossi2021vehicle},
SeqGAN \cite{yu2017seqgan},
MoveSim \cite{feng2020learning},
DiffTraj \cite{zhu2023difftraj},
ControlTraj \cite{zhu2024controltraj}.

\subsubsection{Metrics}
We evaluate the synthetic data quality from both \textit{fidelity} and \textit{utility} perspectives.
Following common practice~\cite{zhu2023difftraj,feng2020learning}, we evaluate data fidelity by computing the Jensen-Shannon Divergence (JSD)~\cite{JSD} between the distributions of synthetic and real data across four metrics from both spatial and temporal aspects: 
1)~\textbf{Distance}: the sum of distance between consecutive check-ins; 
2)~\textbf{Radius}: the radius of gyration, which measures the trajectory's spatial range; 
3)~\textbf{Interval}: the time interval between consecutive check-ins; 
4)~\textbf{Length}: the total number of points in the trajectory.
A lower JSD score indicates a higher fidelity of the synthetic data.
We evaluate the utility of synthetic data by conducting data-driven downstream applications. 

\subsection{Overall Performance}

\subsubsection{Fidelity Evaluation} 
The comprehensive quantitative results are presented in Table~\ref{tab:fidelity}.
We found that traditional rule-based methods (SMM, TimeGEO, Hawkes) generally fall short, as their reliance on fixed statistical assumptions prevents them from capturing the complex, long-term behaviors from POI check-in trajectories.
Designed for spatially continuous and temporally regular trajectory data generation, DiffTraj shows competitive performance on spatial metrics on dense urban datasets like FS-NYC, but performs poorly on the length metric due to the lack of awareness of discrete POI locations.
Although the ControlTraj can model the connection between POIs as road segments, its incompatibility with temporally irregular data degrades the LBSN check-in trajectory generation quality.
In contrast, our proposed model, GeoGen, significantly outperforms baselines across the four diverse datasets, highlighting its robustness and generalizability to generate high-fidelity LBSN trajectory data.
For instance, on the FS-TKY dataset, GeoGen improves the second-best method by over 69\% 
in the Distance metric and 55\% in the Radius metric.
Similarly, it achieves a 38\% improvement in Length on the GW-STO dataset. 

The Cumulative Distribution Function (CDF) curves in Figure~\ref{fig:fidelity} for FS-TKY show that GeoGen's distributions most closely align with the original data, confirming its effectiveness in capturing realistic LBSN check-in behavior patterns.

\subsubsection{Utility Evaluation}
As shown in Table~\ref{tab:utility}, we evaluate the utility of the synthetic data through a next check-in prediction task~\cite{yuan2023spatio}, where a prediction model is trained on synthetic data and tested on real-world data. 
Root Mean Square Error (RMSE) and Euclidean Distance (ED) are used to measure the prediction performance from the temporal dimension and spatial dimension, respectively. 
Our results demonstrate that GeoGen achieves the closest performance to the real data across both temporal and spatial metrics using FS-NYC and FS-TKY datasets, indicating its strong ability to preserve data utility.
MoveSim performs better than SeqGAN and ControlTraj in the spatial dimension due to its explicit consideration of spatial structures such as distances between locations, while SeqGAN and ControlTraj treat locations as discrete categorical variables.
Our proposed \N\ consistently achieves the best performance across all metrics and datasets, thanks to its ability to effectively model spatially discrete and temporally irregular patterns of LBSN check-in trajectories.
\begin{table}[th] \footnotesize
\renewcommand\arraystretch{1.1}
\centering
\begin{tabular}{|c|c|c|c|c|}
\hline
\multicolumn{1}{|c|}{\textbf{Datasets}} & \multicolumn{2}{c|}{\textbf{FS-NYC}} & \multicolumn{2}{c|}{\textbf{FS-TKY}} \\ \hline
\diagbox{\textbf{Methods}}{\textbf{Metrics}} & RMSE  & ED & RMSE & ED \\ \hline
\textbf{Real} & 0.187 & 7.01 & 0.235 & 13.70 \\  \hline
\textbf{SeqGAN} & 0.426 & 16.34 & 0.689 & 23.12 \\ 
\textbf{MoveSim} & 0.371 & 10.28 & 0.671 & 16.90 \\ 
\textbf{DiffTraj} & 0.424 & 15.27 & 0.728 & 18.28\\ 
\textbf{ControlTraj} & 0.321 & 14.29 & 0.491 & 16.28 \\ \hline
\textbf{GeoGen} & \textbf{0.225} & \textbf{9.21} & \textbf{0.352} & \textbf{15.82} \\ \hline
\end{tabular}%
\caption{Synthetic data utility for next check-in prediction. } 
\label{tab:utility}
\end{table}

\subsection{In-depth Analysis}
\subsubsection{Ablation Study} 
To validate our S$^2$TDiff in stage 1, we benchmark it against two other strategies: 
a) replication (S$^2$TDiff-Rep): This method fills empty time slots with the coordinates of the last known check-in.
b) fixed point (S$^2$TDiff-Fix): This method assigns a predetermined central coordinate to all empty time slots.
We also build a variant S$^2$TDiff w/o S$^2$G where the core module S$^2$G Attention module is removed to test its effectiveness.  
For Coarse2FineNet in the second stage, we build a variant C2F w/o DCF by removing the dynamic context fusion component. 
The results on the FS-NYC dataset are shown in Table~\ref{tab:ablation}.
We found that interpolation can provide richer spatio-temporal patterns and underlying physical constraints compared with replication and fix point latent movement sequence reconstruction strategies.
Meanwhile, the performance decreased in both spatial and temporal domains without the S$^2$G Attention module, also proving its effectiveness compared with pure convolutions in the U-Net. 
In Coarse2FineNet, removing the dynamic context fusion and POI embedding (C2F w/o DCF) causes a dramatic performance decrease due to the lack of informative supervision for the learning of the discrete POI embeddings. 
\begin{table}[h]
\centering
\footnotesize
\renewcommand{\arraystretch}{1.1}
\resizebox{\linewidth}{!}{%
\begin{tabular}{|l|c|c|c|c|}
\hline
\diagbox{\textbf{Methods}}{\textbf{Metrics}} & \textbf{Distance} & \textbf{Radius} & \textbf{Interval} & \textbf{Length} \\
\hline
$\text{S}^2\text{TDiff-Rep}$ & 0.084 & 0.071 & 0.074 & 0.128 \\
$\text{S}^2\text{TDiff-Fix}$ & 0.097 & 0.073 & 0.079 & 0.130 \\
$\text{S}^2\text{TDiff w/o S}^2\text{G}$ & 0.087 & 0.069 & 0.080 & 0.135 \\
\text{C2F w/o DCF} & 0.119 & 0.138 & 0.096 & 0.169 \\ \hline
Complete GeoGen & 0.079 & 0.065 & 0.067 & 0.127 \\
\hline
\end{tabular}
}
\caption{Performance comparison across different GeoGen variants on FS-NYC dataset.}
\label{tab:ablation}
\end{table}

\subsubsection{Impact of Granularity on Model Performance and Efficiency}
We investigate the impact of 
time intervals $I$.
As shown in Table~\ref{tab:interval}, 
we found that smaller intervals (e.g., 2-4 hours) yield better temporal fidelity (Interval), whereas spatial fidelity metrics (Distance and Radius) are optimized at a larger 6-hour interval.
Increasing the interval from 2 to 6 hours reduces Stage 1 memory consumption from $2.36$~GB to $0.35$~GB and increases throughput by more than fourfold (from $11,322$ to $50,540$ samples/sec). 
In the second stage, the computational cost of Coarse2FineNet remains relatively stable across different intervals.
Consequently, a moderate $I$ strikes a balance between generation quality and efficiency.

\begin{table}[th]
\renewcommand\arraystretch{1.1}
\centering
\resizebox{\columnwidth}{!}{%
\begin{tabular}{|c|c|c|c|c|c|c|c|}
\hline
\diagbox{\textbf{Metric}}{\textbf{I}} & 2 hours & 3 hours  & 4 hours  & 6 hours & 8 hours & 12 hours\\ \hline
Distance   & 0.454 & 0.183 & 0.136 & \textbf{0.073} & \underline{0.089} & 0.080\\ 
Radius   & 0.402 & 0.250 & 0.197 & \textbf{0.154} & \underline{0.153} & 0.154\\
Interval & \textbf{0.292} & \textbf{0.292} & \textbf{0.292} & 0.315 & 0.342 & 0.444\\
Length  & 0.318 & 0.137 & \textbf{0.080} & \underline{0.115} & 0.134 & 0.131\\
\hline
S1 Memory (GB) & 2.36 & 1.14 & 0.69 & 0.35 & 0.21 & 0.12\\
S2 Memory (GB) & 17.55  & 9.78 & 10.10 & 10.68 & 3.62 & 3.92\\
S1 Throughput  & 11322 & 20221 & 30058 & 50540 & 73794 & 119932\\
S2 Throughput  & 39 & 41 & 44 & 45 & 49 & 49\\\hline
\end{tabular}%
}
\caption{Results under different coarse-grained interval $I$ selection on GW-STO dataset with batch size 256 for S$^2$TDiff  (S1) and 16 for Coarse2FineNet (S2). 
}
\label{tab:interval}
\end{table}



\section{Related Work}

\subsection{LBSN Data Mining}


In recent years, LBSN data have attracted much interest from both research and industry communities~\cite {lbsn_1,lbsn_survey} due to their high value for many practical applications such as POI recommendations~\cite{next_loc_prediction}, user behavior analysis~\cite{lbsn_survey}, and advertising~\cite{jeon2021lightmove}. 
Yang et al.~\cite{next_poi_rec_2} design a self-explainable algorithm to recommend the next visiting POI with high accuracy based on LBSN data.
Han et al.~\cite{han2021will} leverage LBSN data to understand human activities during COVID-19 for better pandemic control strategies. 
However, given the potential privacy concerns and high collection costs, it becomes highly challenging for researchers to access large-scale LBSN data~\cite{poi_privacy,kunlin}. 
Hence, in this study, we focus on generating large-scale synthetic LBSN data that shares similar data characteristics with real data and also has high utility for downstream applications. 

\subsection{Trajectory Data Generation}
The recent advances in generative models provide a great opportunity for synthetic trajectory data generation. These models (e.g., Generative Adversarial Network (GAN) \cite{rao2020lstm, liu2018trajgans, cao2021generating} and diffusion models \cite{zhu2023difftraj}) utilize the power of deep neural networks to learn and replicate the underlying spatio-temporal distributions of real trajectory data. Once trained, these models generate synthetic trajectories by sampling from the learned distributions. 
For example, Feng et al. \cite{feng2020learning} propose a GAN-based framework, MoveSim, with the introduction of prior knowledge for human mobility trajectory generation. 
MoveSim \cite{feng2020learning} generates the next location based on the last point in the partially generated sequence and iteratively generates the rest trajectory points. Zhu et al. \cite{zhu2023difftraj, zhu2023synmob, zhu2024controltraj} propose a pioneering work by adopting the diffusion model to generate continuous and road network-constrained trajectories with a fixed length and time interval between two trajectory points. 
Different from existing works that usually focus on generating trajectories in continuous spatial spaces with fixed time intervals and fixed lengths, our work aims to generate spatially discrete, temporally irregular LBSN check-in trajectories.

\section{Conclusion}
In this paper, we propose GeoGen, a novel two-stage coarse-to-fine framework for the generation of synthetic LBSN check-in trajectories that are characterized by spatial discreteness and temporal irregularity. In the first stage, we propose 
a Sparsity-aware Spatio-temporal Diffusion model (S$^2$TDiff) with an efficient Spatially-Aware Sparsely-Gated U-Net (SASG-UNet) to learn multi-scale behavioral patterns. In the second stage, we design
Coarse2FineNet, a Transformer-based Seq2Seq architecture equipped with a Dynamic Context Fusion mechanism in the encoder and a multi-task hybrid-head decoder to generate accurate next POI location and its fine-grained timestamp. We extensively evaluate our GeoGen with four datasets, and results show that \N outperforms state-of-the-art models from both data fidelity and utility perspectives, e.g., it increases 69\% and 55\% on distance and radius metrics on the FS-TKY dataset, and enhances the next check-in prediction task.


\section{Acknowledgments}
We sincerely thank all anonymous reviewers for their insightful comments and valuable suggestions.
This work is partially supported by the National Science Foundation (NSF) Awards 2411151, 2411152, 2411153, National Artificial Intelligence Research Resource (NAIRR)
240332, and FSU/AWS Computer Support Seed Fund. 

\bibliography{main.bbl}

\end{document}